\pdfoutput=1

\documentclass[11pt]{article}

\usepackage[preprint]{acl}

\usepackage{times}
\usepackage{latexsym}

\usepackage[T1]{fontenc}

\usepackage[utf8]{inputenc}

\usepackage{microtype}

\usepackage{inconsolata}

\usepackage{graphicx}
\usepackage[breaklinks]{hyperref}
\usepackage{xcolor}
\usepackage{amsmath}
\usepackage{algorithm}
\usepackage{algorithmic}
\usepackage{enumitem}
\usepackage{multirow}
\usepackage{tabularx}
\usepackage{array}
\usepackage{booktabs}
\usepackage{siunitx}
\usepackage{appendix}
\usepackage{float}
\usepackage{xspace}
\usepackage{bbm}

\title{LLM-based Privacy Data Augmentation Guided by Knowledge Distillation with a Distribution Tutor for Medical Text Classification}

\author{{\bf Yiping Song$^{1,\ast}$, Juhua Zhang$^{2,}$\thanks{Equal Contribution.}, Zhiliang Tian$^{2,}$\thanks{Corresponding Author.},}\\  {\bf Yuxin Yang$^{2}$, Minlie Huang$^{3}$, Dongsheng Li$^{2}$} \\
$^{1}$College of Science, National University of Defense Technology, Hunan, China\\$^{2}$College of Computer, National University of Defense Technology, Hunan, China \\$^{3}$The CoAI Group, DCST, BNRist, Tsinghua University, Beijing 100084, China \\ 
\texttt{\{songyiping, zhangjuhua23, tianzhiliang,}\\
\texttt{yangyuxin21a, dsli\}@nudt.edu.cn, aihuang@tsinghua.edu.cn}
}

\begin{document}
\maketitle
\begin{abstract}

As sufficient data are not always publically accessible for model training, researchers exploit limited data with advanced learning algorithms or expand the dataset via data augmentation (DA). Conducting DA in private domain requires private protection approaches (i.e. anonymization and perturbation), but those methods cannot provide protection guarantees. Differential privacy (DP) learning methods theoretically bound the protection but are not skilled at generating pseudo text samples with large models. In this paper, we transfer DP-based pseudo sample generation task to DP-based generated samples discrimination task, where we propose a DP-based DA method\footnote{Code is available at \url{https://anonymous.4open.science/r/DP_DA-1BF7}} with a LLM and a DP-based discriminator for text classification on private domains. We construct a knowledge distillation model as the DP-based discriminator: teacher models, accessing private data, teaches students how to select private samples with calibrated noise to achieve DP. To constrain the distribution of DA's generation, we propose a DP-based tutor that models the noised private distribution and controls samples' generation with a low privacy cost. We theoretically analyze our model's privacy protection and empirically verify our model.
\end{abstract}

\section{Introduction}
The ability of deep text classification models mainly derives from large-scale training data and recent large language models (LLMs) particularly benefit from big data. Many domains (e.g. medicine \cite{qing2019novel,li2021hybrid}) have only limited amount of public data and large private data. It is risky to release private data to model training since models probably memorize details in those data and unintentionally output their sensitive information \cite{carlini2019secret,carlini2021extracting}.

Researchers fully exploit limited public data and avoid accessing private data to ensure data security \cite{fei2006one}, which achieves learning on small data via meta-learning \cite{maml_2017icml} or active learning \cite{konyushkova2017learning}. Further, some researchers expand the public dataset with data augmentation (DA) \cite{wei2019eda}; another promising DA strategy is to synthesize private samples conditioned on privacy data while ensuring the sensitive information in private data are well-protected \cite{yue2022synthetic}. 

Synthesizing private text requires to capture original private data distributions by accessing the data. Accessing them should be under the guarantee of privacy protections and thus we need privacy protection on text generation. Researchers attempt anonymization methods to mask selected sensitive text span \cite{secret,shi2021selective}, randomly noise the input tokens \cite{feyisetan2020privacy}, or apply regularization to avoid overfitting training data \cite{secret}. However, the masking, nosing, or regularization mechanism hardly ensures (almost) all personal information is well-protected. Differential privacy (DP) \cite{dwork2006calibrating} provides provable guarantees against the identification of individual information in datasets. Deep generative models with DP ensure the existence of a specific sample (with private information) cannot be detected \cite{li2021large}.

Noisy-SGD ~\citep{song2013stochastic,dpsgd2016deep} is a practical DP algorithm for deep learning models, including text generation models \cite{dpsgd_nlp_2020differentially}, which adds calibrated noise on model gradients to satisfy DP. However, the advantage of NoisySGD is weaken as the model becomes larger \citep{bassily2014private,yu2020not} and NoisySGD requires a per-example gradient clip resulting in non-convergence and system overheads \citep{knn,bu2021convergence}. Another categories of DP learning methods, PATE \citep{pate2017}, acquires private information from teacher models learned on private data to a student and adds noises on teachers' outputs. PATE's privacy cost comes from teachers' outputs instead of the whole model's gradients in Noisy-SGD \citep{li2021large}. Hence, PATE's required noise does not scale with model size and PATE is promising of working on large models (e.g. BERT \cite{bert2019naacl_devlin}, Llama ~\cite{touvron2023llama}, and GPT-4 ~\cite{achiam2023gpt}).

DP with PATE on text generation still suffers from a sequential multiple tasks with large candidate space, which extremely (linearly) increases the noise scale \cite{tian2022seqpate}. We argue that we can transfer a DP-based generation task to a DP-based discrimination task to avoid complexities in DP text generation models. Particularly, we employ LLMs' generation ability $P_{LLM}(x)$ to generate public samples $x$ and construct a DP-based discriminator $P_{discri}(\cdot|x)$ to select synthesized samples $x$ fits for private domain $c$, synthesizing private samples via $P_{syn}(x|\cdot) = P_{discri}(\cdot|x) * P_{LLM}(x)$.

In this paper, we propose a DP-based data augmentation (DA) paradigm with a LLM and a DP-based discriminator to generate samples for private text classification, where the discriminator selects LLM-generated samples likely to belong to private domain as our synthesized samples. Specifically, the discriminator achieves DP via knowledge distillation (KD), where multiple teacher models access disjoint and unique private sets to learn private discriminators; a student learns from noised aggregated teacher outputs to achieve DP. To control the distribution in DA's generated samples, we propose a DP-based distribution tutor that captures the distribution of private data. 
In DP, querying privacy is expensive (student querying teacher causes a certain privacy cost), the tutor carries less sensitive information than teachers and helps the student with a low privacy cost (See §\ref{sec:privacy_analyses}).
We further provide theoretical analyses and empirical results to verify our methods. 

Our contributions are as follows:
(1) We construct a DP-based DA with LLMs that synthetizes (almost infinite) samples while bounding the privacy leakage.
(2) We propose a DP-based tutor for teacher-student frameworks to teach some less sensitive data with a low privacy cost.
(3) We excel strong DP-based baselines on text classification in private domains.

\section{Related Work}
\subsection{Privacy Protection in Text Classification}

Initial privacy protection techniques predominantly focused on data anonymization \cite{maeda2016fast,suzuki2018k}, For example, de-identification techniques \cite{garfinkel2015identification} such as removing, replacing, or encrypting sensitive information in data can reduce the risk of privacy leaks. Data perturbation techniques \cite{johnson2013privacy} protect user privacy by incorporating random noise into the data. 
Nevertheless, straightforward data anonymization measures may difficult to effectively deal with privacy leakage challenges \cite{rocher2019estimating}. Presently, federated learning 
\cite{mcmahan2017communication,deng2022secure} and DP \cite{dwork2006calibrating} emerge as the two principal methodologies in the domain of privacy protection. Federated learning strategies can prevent privacy leaks caused by untrustworthy servers. DP aims to prevent attackers from extracting sensitive information from the training dataset \cite{
carlini2021extracting}, offering a quantifiable privacy protection mechanism. With its robust theoretical foundation and broad applicability\cite{dwork2014algorithmic}, differential privacy is widely acknowledged as the standard practice in the field of privacy protection. Independently and concurrently with our work, \cite{wu2023privacy} and \cite{duan2023flocks} studied ICL with DP guarantees for text classification tasks. Our proposed method predominantly embraces the privacy protection principles of differential privacy.

\subsection{Data Augmentation (DA) in Text Classification}
Various NLP data augmentation (DA) techniques have been developed, such as Back-Translation \cite{kobayashi2018contextual}, EDA \cite{wei2019eda}, and AEDA \cite{karimi2021aeda}. These methods primarily focus on modifying the original input, which limits the diversity of the generated samples. In response, \cite{szegedy2016rethinking} initially explore a interpolation-based methods (i.e., mixup) in computer vision. Subsequently, \cite{guo2019augmenting} combine the mixup technique with CNNs and LSTMs for text applications.There are also many studies that choose different strategies to improve the mixup technique \cite{chen2020local,zhang2020seqmix}.

Moreover, some researchers use pre-trained language models (PLMs) for data for data augmentation. \cite{kumar2020data} provide a straightforward and effective method for conditional PLM by prepending class labels to text sequences. \cite{hu2019learning} utilize reinforcement learning with a conditional language model that performs by appending the correct label to the input sequence during training. Further, an increasing number of scholars have started to utilize adversarial learning methods to generate augmented samples, such as BERT-Attack \cite{li2020bert}, G-DAUG$^C$ \cite{yang2020generative}.

To reduce the negative impact of low-quality augmentation samples on model performance, some research focus on sample selection. For example, 
\cite{cao2021uncertainty} propose UAST framework to quantify model uncertainty for selecting pseudo-labeled samples. \cite{lin2023choose} focus more on the combination of sample selection and data enhancement strategies, and introduce a self-training selection framework to select high-quality samples from the data augmentation. 
Different from the above methods, we aim to achieve DA to synthetic private data while ensuring the private information from the private dataset.

\subsection{DP for Deep Learning Models}
Some researchers employ DP to protect the privacy of empirical risk minimization classifiers \cite{chaudhuri2011differentially} and SVM \cite{rubinstein2009learning}. Following \cite{song2013stochastic}, NoisySGD introduces noise into gradients to achieve DP for deep learning models, including DP-SGD \cite{song2013stochastic,bassily2014private,bu2023convergence} and DP-Adam \cite{dpsgd2016deep,kingma2014adam}. The use of DP-SGD for large-scale pre-training of BERT has been shown to achieve comparable masked language modeling performance to non-private BERT \cite{anil2021large}, but with a privacy budget is 100 or higher. Recent studies have demonstrated that even under more stringent privacy constraints, generative and discriminative language models can achieve high performance across various tasks by appropriately selecting hyperparameters and fine-tuning objectives aligned with the pre-training process \cite{li2021large}. Additionally, \cite{li2021large} apply ghost clipping to pre-trained language models using NoisySGD, reducing memory usage. \cite{he2022exploring} leverage group clipping with adaptive clipping thresholds, privately fine-tuning GPT-3 with 1.75 trillion parameters. PATE \citep{pate2017} is another type of DP learning algorithm, transferring knowledge from teacher models trained on private sets with noises to a student model. The privacy cost of PATE arises from knowledge distillation rather than the gradient of the entire model. With this advantage, PATE has enormous potential in adapting to large models. Moreover, PATE is designed for classification tasks and is suitable for our goal of training a DP-based discriminator.

\begin{figure*}[t]
    \centering
    \includegraphics[width=1\textwidth]{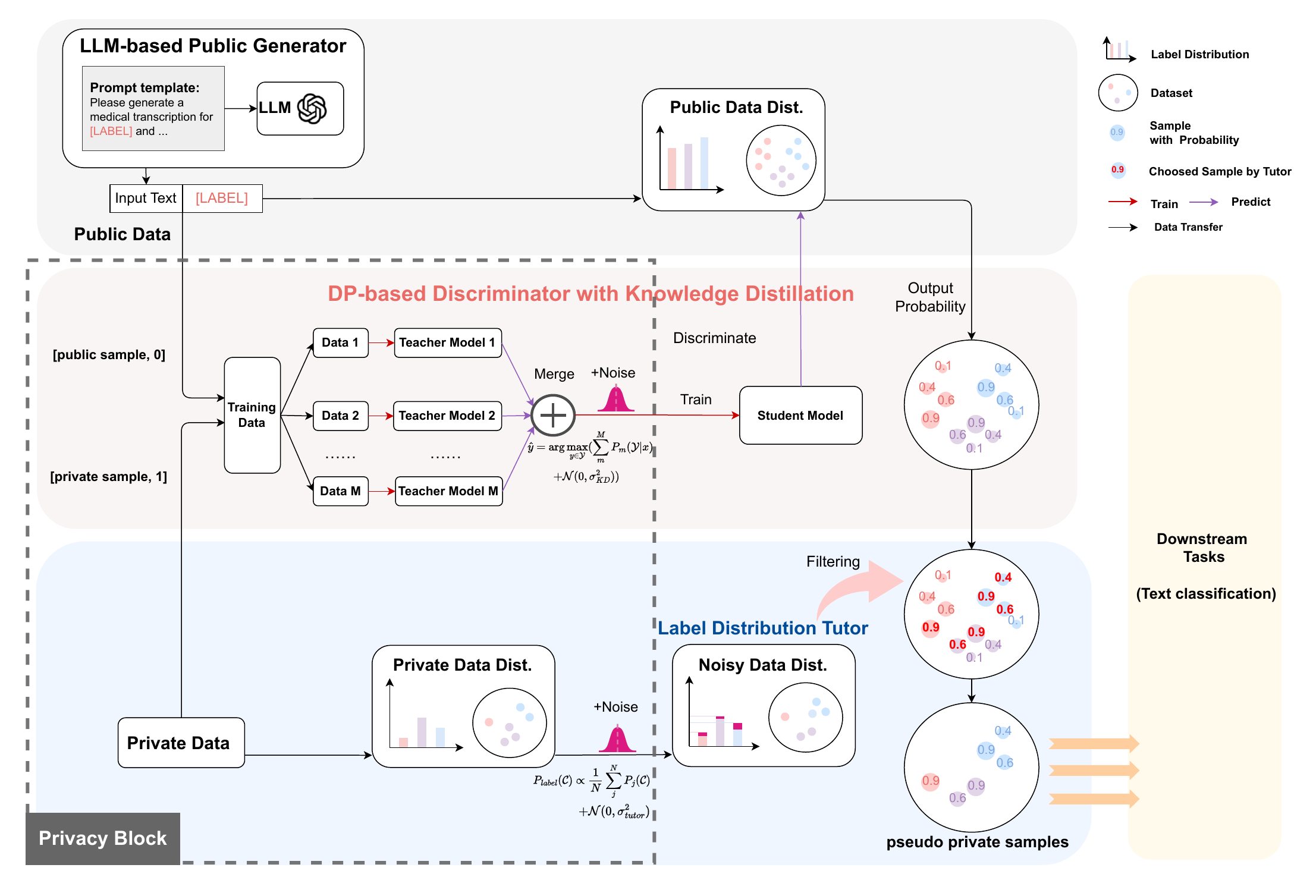}
    \caption{Overview of our framework. It mainly contains three components, LLM-based Public Generator (grey block) generates public data. DP-based Discriminator with Knowledge Distillation (pink block) discriminates public data and obtains a probability similar to private data. The Label Distribution Tutor (blue block) selects a subset with the highest probabilities of samples matching the noise label distribution. The gray dotted box is the privacy block.}
    \label{fig1}
\end{figure*}

\section{Methods}
\subsection{Overview}
\label{sec:overview}
Our model consists of four parts (Fig.~\ref{fig1}).
\paragraph{\textbf{LLM-based Public Generator}} (§\ref{sec:generator}) synthetic public textual input for the specific label on the text classification task.
\paragraph{\textbf{DP-based Discriminator with Knowledge Distillation}} (§\ref{sec:discriminator}) learns to discriminate whether a sample is (likely) from public domain or private domain, which satisfy DP privacy guarantee. It filters the samples from the generator (§\ref{sec:generator}) to obtain new samples for privacy domain.
\paragraph{\textbf{Label Distribution Tutor}} (§\ref{sec:tutor}) leads the generated pseudo samples to follow the distribution of private data under DP guarantee, which also filter the generator §\ref{sec:generator}'s output.
\paragraph{\textbf{Private Data Augmentation}} (§\ref{sec:framework}) uses the generator to obtain candidate samples and uses the discriminator (§\ref{sec:discriminator}) and the tutor (§\ref{sec:tutor}) to filter the candidates to get data augmentation samples.

\subsection{Task Definition}
\label{sec:task_definition}
Given a private text classification corpora with sensitive information, our task aims to train a text classifier with a certain level of theoretical privacy guarantee, which protects all training samples with its label $<x, c>$ from being detected (protecting the existence of specific training sample from being identified by any detectors via any detecting methods).


Note that our task allows the model using data augmentation (DA) method to generate pseudo private samples for training. However, the data augmentation model also ought to ensure the privacy guarantee mentioned above.

\subsection{LLM-based Public Generator}
\label{sec:generator}
We employ a LLM (i.e. GPT-3.5) to generate input texts for each output label (on classification task). Even if the GPT-3.5 is trained on the public domain, we design a prompt text to induce the LLM to attempt to generate private samples, where the prompt follows this template: ``You are a professional medical transcriber. Please generate a medical transcription for [LABEL] and do not reveal the patient's name. The text length of medical transcription is approximately 400 words and at least 200 words.'' In the above template, ``[LABEL]'' denotes the label $c$ of classifications (i.e. medical specialty). In this way, LLM generates the input text $x$ with the label $c$ to compose a complete pseudo sample $<x, c>$.



 

\subsection{DP-based Discriminator with Knowledge Distillation}
\label{sec:discriminator}
We propose a DP-based discriminator to check if the pseudo samples fit for the private distribution thus are capable of acting as private samples.
Inspired by \cite{pate2017}, we construct a teacher-student framework with knowledge distillation. This model consists of multiple teacher models and a student model. Teachers are allowed to access the private data and the student can only access the noised teachers' outputs. 

\paragraph{\textbf{Teacher Models.}} We train multiple teacher models on multiple disjoint datasets, where the private samples act as the positive sample and the public samples (generated by §\ref{sec:generator}) as the negative sample. The teachers learn to judge whether the samples from private set or public set. All teachers follow the structure and the initial parameters of a pre-trained model.

To satisfy DP, we carefully process the teachers' data and train teachers with two strategies. First, the private training data need to contain only unique samples (private sample duplicates should be removed). The reason is that DP prevents the unknown detectors from identification on each occurrence \cite{dwork2006calibrating}.  Duplicated samples with $N$ occurrences increase the private loss and thus decrease the bound of privacy protection (in terms of $\varepsilon$) \cite{dwork2014algorithmic}. To maintain the bound, the noise scale has to increase $N$ times \cite{dpsgd_nlp_2020differentially} (or $\log(N)$ times s.t. advanced DP \cite{li2021large}) which extremely harms for the performance. 

Second, we equally divide the shuffled private data into $M$ disjoint sets for $M$ teachers and train each teacher on each set separately. In this way, the existence of any specific sample affects only one teacher's output, which bounds the sensitivity of the teachers' noisy distribution \cite{pate2017,boenisch2023individualized} (See detailed analyses in §\ref{sec:privacy_analyses}). The equally divide (in terms of the sample number) and shuffled assignment ensure the balance among all teachers and the training performance.
    
\paragraph{\textbf{Student Model.}} The student model is a discriminator cannot access raw private data and learn from teachers with those steps:
(1) \texttt{Merging}. For a given sample $x$, we merge teachers' predictions $P_{m}(\mathcal{Y}|x)$, where $\mathcal{Y}=\{0,1\}$ indicates whether the sample $x$ derives from private or public set. (2) \texttt{Noising}. According to DP theory \cite{dpbook014algorithmic}, we add calibrated noise on the teacher's predictions to obtain $\hat{y}$ as Eq.~\ref{eq:aggregate}, denoting teachers' estimated judgment on $x$.
\begin{align}
\label{eq:aggregate}
    \hat{y} &= f_{tea\_agg}(\mathcal{Y}|x) \\
    &= \arg \max_{y \in \mathcal{Y}} (\sum^M_{m} P_{m}(\mathcal{Y}|x) + \mathcal{N}(0, \sigma_{KD}^2)) \nonumber
\end{align}
, where the first term is the actual teacher aggregated outputs and $M$ indicates the teacher number; the second term represents the calibrated Gaussian noise \cite{bu2020deep} and $\sigma_{teacher}$ controls the degree of adding noise (i.e. degree of privacy protection).
(3) \texttt{Teaching}. We employ the noisy outputs $\hat{Y}$ to act as pseudo labels, which teaches the student how to discriminate the privacy samples. For each sample $x$, the teaching follows Cross-Entropy loss as Eq.~\ref{eq:ce_loss}.

\begin{align}
\label{eq:ce_loss}
    \mathcal{L}_{KD} &= \text{CE}(f_{tea\_agg}(\mathcal{C}|x), P_{student}(\mathcal{C}|x)) \\
    &= - \sum_m^M \mathbbm{I}(c_m = \hat{c}) \log P_{student}(c_m | x) \nonumber
\end{align}
, where $\mathbbm{I}$ is a indicator function. The above mechanism ensures the student satisfies DP as~\cite{pate2017} (Analyses in §\ref{sec:privacy_analyses}). 


\subsection{Label Distribution Tutor}
\label{sec:tutor}
We propose a tutor based on the above teacher-student framework (§\ref{sec:discriminator}), which avoids directly access the pure private samples but accesses only a small amount of privacy information so as to maintain the bound of protections. The tutor aims to carry the label distribution of the private samples, as the label distribution is critical in generating augmented data.

The tutor follows a statistic way to collect the noised label distribution of all private samples $P_{label}(\mathcal{C})$ with a very small privacy cost. The tutor first obtains the actual label distribution by counting the total $N$ training samples as the first term of Eq.~\ref{eq:tutor}. Then, it imports calibrated Gaussian noise on the label distribution as the second term of Eq.~\ref{eq:tutor} \footnote{Mathematically, the noisy aggregated output of Eq.~\ref{eq:tutor} is expected to follow probability distribution (range from 0 to 1) as the mean of noises is 0. But some output values may be occasionally out of [0,1]. If so, we follow \citet{tian2022seqpate} to re-normalize the out-of-bound value to 0 or 1. Practically, we observed (in Fig. 2) being out-of-bound is extremely rare since $N$ is large (3$k$) and the first term dominates Eq.~\ref{eq:tutor} as \cite{tian2022seqpate}.}. The above mechanism ensures the tutor satisfy DP (Analyses in §\ref{sec:privacy_analyses}).

\begin{equation}
\label{eq:tutor}
P_{label}(\mathcal{C}) \propto \frac{1}{N} \sum_{j}^{N} P_{j}(\mathcal{C}) + \mathcal{N}(0, \sigma_{tutor}^2).
\end{equation}

\subsection{Private Data Augmentation}
\label{sec:framework}

We propose a novel data augmentation (DA) framework to generate samples for private domain while protecting the privacy. The framework involves: (1) a public LLM-based generator to obtain a plenty of candidate samples (§\ref{sec:discriminator}). (2)
a DP-based discriminator to select LLM's generated samples similar to private samples, and (3) a tutor to filter generated samples to ensure the label distributions fitting for the private data.

The idea is (1) taking advantage of strong generation ability of LLMs to obtain high quality data. (2) accessing private data causes privacy cost but accessing privacy through the student (i.e. discriminator) and tutor satisfying DP would not brings in additional loss. Hence, we can ``infinitely'' call LLMs, student, and tutor to achieve DA while bounding the privacy protection. 



\section{Privacy Analyses of our Method}
\label{sec:privacy_analyses}



\paragraph{\textbf{Lemma 4 Analytical Gaussian mechanism.}} \citep{balle2018improving}
\label{lm:ana_gau}
For a query $h: \mathcal{X}^n \rightarrow \mathcal{Y}^d$ over a dataset $\mathcal{D}$, the randomized algorithm outputting $h(\mathcal{D}) + Z \;\; s.t. \;\; Z \sim \mathcal{N}(0, \sigma^2 I_d)$ satisfies $(\varepsilon,\delta(\varepsilon)$)-DP for all $\varepsilon\geq 0$ and $\delta(\varepsilon) = \Phi(\frac{\Delta}{2\sigma}-\frac{\varepsilon\sigma}{\Delta}) - e^\varepsilon \Phi(-\frac{\Delta}{2\sigma}- \frac{\varepsilon\sigma}{\Delta})$, 
where $\Delta = \max_{\mathcal{D} \sim \mathcal{D}^\prime} \| h(\mathcal{D}) - h(\mathcal{D}^\prime) \|_2$ is L2 sensitivity of $h$ and $\Phi$ is CDF function of $\mathcal{N}(0,1)$.

\begin{table*}[h!]
\centering
\setlength{\tabcolsep}{3.3pt}
\begin{tabular}{cccccccccc}
\toprule
\multicolumn{2}{c}{\multirow{2}{*}{Method}}         & \multicolumn{4}{c}{3750 training samples}                                         & \multicolumn{4}{c}{6000 training samples}                                         \\ \cline{3-10}
\multicolumn{2}{c}{} & P     & R         & F1       & Acc       & P     & R         & F1       & Acc     \\
\midrule
\multirow{2}{*}{Non-DP}              & Private      & 0.224         & 0.367          & 0.240           & 0.367          & -              & -              & -             & -              \\
 & DA w/ Public & 0.280          & 0.347          & 0.280           & 0.348          & 0.313          & 0.347          & 0.287         & 0.348                  \\
\midrule
\multirow{2}{*}{DP($\varepsilon=4$)}       & DP-SGD       & 0.130          & 0.310           & 0.170           & 0.314          & -              & -              & -             & -             \\
 & Ghost        & 0.143         & 0.289          & 0.166          & 0.291          & -              & -              & -             & -              \\
\midrule
\multirow{3}{*}{DA w/ DP($\varepsilon=4$)} & DA w/ DP-SGD & 0.290          & 0.350           & 0.277          & 0.352          & 0.283          & 0.350           & 0.280          & 0.352                 \\
 & DA w/ Ghost  & 0.303         & 0.347          & 0.297          & 0.344          & 0.283          & 0.350           & 0.297         & 0.350           \\
 & Ours         & \textbf{0.340} & \textbf{0.373} & \textbf{0.337} & \textbf{0.372} & \textbf{0.353} & \textbf{0.377} & \textbf{0.340} & \textbf{0.376} \\
\bottomrule
\end{tabular}
\caption{Main results comparing all the baselines on two size of datasets on text classification tasks.}
\label{tab:main}
\end{table*}

\paragraph{\textbf{Sensitivity Analysis of Knowledge Distillation (KD).}} We denote the output distribution of $m$-th teacher model is 
$P_{m}(\mathcal{Y}|x)$. The aggregation function $h(\mathcal{D}) = \sum_{m=1}^M P_m(\mathcal{Y}|x)$ is the summation of the output probability over all the teachers. Note that each teacher's data are disjoint (§\ref{sec:discriminator}) and teachers have no duplicate samples, changing one sample only affects one teacher's output. For neighboring datasets $\mathcal{D}$ and $\mathcal{D}^\prime$ differing only one sample, we denote $i$-th teacher is affected by the different sample in $\mathcal{D}$ and $\mathcal{D}^\prime$. The output distributions $P_i(\mathcal{Y}|x)$ and $P^\prime_i(\mathcal{Y}|x)$ (from $\mathcal{D}$ and $\mathcal{D}^\prime$ respectively) are different. The sensitivity $\Delta_{KD}$ in Lemma~\ref{lm:ana_gau} is (See deductions in App.~\ref{sec:sensitivity_kd}),

\begin{align}
\label{eq:sensitivity_kd}
    \Delta_{KD} &= \| h(\mathcal{D}) - h(\mathcal{D'}) \|_2 \\
    &\le \| P_i(\mathcal{Y}|x) - P_i^\prime(\mathcal{Y}|x) \|_2 \le \sqrt{2}. \nonumber
\end{align}

\paragraph{\textbf{Sensitivity Analysis of Tutor.}}
In Eq.~\ref{eq:tutor}, each sample $j$'s distribution $P_{j}(\mathcal{C}) = \{P_j(c_i)\}_{i=1}^{|\mathcal{C}|}$ is the binary-valued probability distribution, its value on the label is 1 while the value on all other categories is 0. $P_j(c_i) = \mathbbm{I}(c_i \text{ is } x_j \text{'s label})$, where $\mathbbm{I}$ is the indicator function. The function   $g(\mathcal{D}) = \sum_{j}^{N} P_{j}(\mathcal{C})$ is the distribution on whole datasets. For neighboring dataset $\mathcal{D}$ and $\mathcal{D}^\prime$, the different sample $l$ in $\mathcal{D}$ and $\mathcal{D}^\prime$ affects two terms:  $P_{l}(\mathcal{C})$ in $g(\mathcal{D})$ and $P^\prime_{l}(\mathcal{C})$ in $g(\mathcal{D}^\prime)$. The sensitivity $\Delta_{tutor}$ in Lemma~\ref{lm:ana_gau} is (See deductions in App.~\ref{app:sensitivity_tutor}),

\begin{align}
\label{eq:sensitivity_tutor}
    \Delta_{tutor} &= \| g(\mathcal{D}) - g(\mathcal{D'}) \|_2 \\
    &\le \| P_l(\mathcal{C}) - P_l^\prime(\mathcal{C}) \|_2 \le \sqrt{2}. \nonumber
\end{align}

\paragraph{\textbf{Composition of KD and Tutor.}}
Our whole DP learning algorithm is actually the combination of KD algorithm (§\ref{sec:discriminator}) and tutor algorithm (§\ref{sec:tutor}). According to  composition theorem for ($\varepsilon$, $\delta$)-DP ~\cite{dwork2006calibrating}, when the KD algorithm $\mathcal{M}_{KD}$ satisfies ($\varepsilon_{KD}$, $\delta_{KD}$)-DP and the tutor algorithm $\mathcal{M}_{tutor}$ satisfies ($\varepsilon_{tutor}$, $\delta_{tutor}$)-DP, then the combination (i.e. our whole algorithm) $\mathcal{M}$ satisfies ($\varepsilon_{KD} + \varepsilon_{tutor}$, $\delta_{KD} + \delta_{tutor}$)-DP. (See deduction in App.~\ref{app:composition}). In summary, adding the noises according to in Lemma~\ref{lm:ana_gau} to KD and the tutor is sufficient to preserve ($\varepsilon_{KD} + \varepsilon_{tutor}$, $\delta_{KD} + \delta_{tutor}$)-DP for the composition of KD and the tutor where the sensitivity is $\sqrt{2}$. 

\section{Experimental Settings}
\paragraph{\textbf{Datasets.}} We evaluate the methods on ``Medical Transcriptions"\footnote{www.kaggle.com/tboyle10/medicaltranscri\-ptions} dataset, which is a dataset in the medical domain with medical transcription samples from  40 various medical specialties. It contains 5k items. Medical data are extremely hard to find due to HIPAA privacy regulations \cite{act1996health}. This dataset was scraped from mtsamples.com.
We performed basic text processing on the data, converting all text to lowercase and removing punctuation. Subsequently, we randomly divided 75\% of the samples for training and 25\% for testing.

\paragraph{\textbf{Evaluation Metrics.}}
Metrics consists of: accuracy (\texttt{Acc}), precision (\texttt{P}), recall (\texttt{R}), and F1-score (\texttt{F1}, the harmonic mean of \texttt{P} and \texttt{R}).

\paragraph{\textbf{Comparing Methods.}}
We used two non-DP methods as the performance upper or lower bound: (1) \texttt{Private} directly trains on private data without protections.
(2) \texttt{DA w/ Public} trains on data synthesized by public GPT-3.5 without accessing private data.

We use DP-based methods as: (1) \texttt{DP-SGD} \cite{dpsgd2016deep} trains on private data based on DP-SGD with noises on gradients. (2) \texttt{Ghost} \cite{li2021large} trains on private data with DP-Adam and ``ghost clipping''.
(3) \texttt{DA w/ DP-SGD} trains DP-SGD on private data to select pseudo DA samples.
(4) \texttt{DA w/ Ghost} trains Ghost on private data to select pseudo DA samples.
(5) \texttt{Ours} denotes our proposed method.
Note that \texttt{DA w/ DP-SGD} and \texttt{DA w/ Ghost} also imitates the noisy label distribution for a fair comparison with \texttt{Ours}.
(See implementation details in App.~\ref{sec:implementation details}).

\section{Experimental Results}

\subsection{Main Results}
Table~\ref{tab:main} presents the  overall performance and we can observe that: 
without data augmentation (DA), \texttt{Private} acts as performance up-bound since it fully accesses the private data without any private protection. There are only 3750 train samples without DA.
The Noisy-SGD methods (i.e. \texttt{DP-SGD} and \texttt{Ghost}) with DA exhibit significant increases to the same methods without DA, which shows the effectiveness of our designed DA framework. Note that we keep the sample number of DA and non-DA methods same for a fair comparison, and DA still works better since DA obtains high quality samples with a less privacy cost.

\texttt{Ours} outperforms other baseline methods in a meaningful range of privacy protection ($\varepsilon$ is 4), even \texttt{Private}. The reason for outperforming \texttt{Private}, which acts as the up-bound, is that our synthesized data have higher quality and diversity than the private data since they are sampled from LLMs, which improves the generalization ability of classification models as training data are not sufficient.
When the training samples selected from a fixed-size synthetic dataset increase from 3750 to 6000, the performance of all methods generally improves. 


\subsection{Ablation Studies}
\begin{table}[t]

\centering
\setlength{\tabcolsep}{3.3pt}
\resizebox{\linewidth}{!}{
\begin{tabular}{ccrrr}
\toprule
\multicolumn{1}{c}{\multirow{2}{*}{Method}}         & \multicolumn{4}{c}{3750 training samples}                                        \\ \cline{2-5}
\multicolumn{1}{c}{} & P     & R         & F1       & Acc     \\
\midrule
Ours  & 0.340                        & 0.373 & 0.337   & 0.372   \\ \hline
Ours $-$ Multi-teacher & 0.280 & 0.297 & 0.213   & 0.298   \\ 
Ours $-$ Gaussian     &0.327                             &0.333        &0.263          &0.335          \\  \hline
Ours $-$ Tutor    & 0.340                        & 0.340   & 0.313   & 0.340   \\ 
Ours $-$ Tutor $+$ Label Dist.    & 0.347                      & 0.377 & 0.340     & 0.377   \\
\bottomrule
\end{tabular}
}
\caption{Ablation studies. ``$+$" or ``$-$" means using or not using the given strategy.}
\label{tab:ablation}
\end{table}
We conducted ablation studies to evaluate the effectiveness of our proposed components. Table \ref{tab:ablation} presents the precision, recall, F1-score, and accuracy of \texttt{Ours}' variants on the downstream task. (1) \texttt{Ours $-$ Multi-teacher}: only use one teacher to train the discriminator instead of multiple teachers, where its subpar performance is due to this teacher completely determines the prediction result when there is only one teacher model, and the noise will directly affect the prediction result and bring a great negative impact.
where its subpar performance is due to when there is only one teacher model, this teacher completely determines the prediction result, and the noise will directly affect the prediction result and bring a great negative impact.
(2) \texttt{Ours$-$Laplace}: use the Laplace mechanism to add noise to the discriminator instead of the Gaussian mechanism. The result demonstrates that the performance of using the Gaussian mechanism in our framework is better than using Laplacian noise. (3) \texttt{Ours $-$ Tutor}: discard the tutor (with the label distribution) and simply select samples uniformly considering the label. The poor performance shows the significance of the proposed tutor and using label distribution in DA. (4) \texttt{Ours $-$ Tutor $+$ Label Dist.}: 
discard the tutor but directly imitate the private label distribution to verify the effectiveness of learning the private label distribution, which is confirmed by its excellent performance. Compared with \texttt{Ours}, it can learn a label distribution closer to real data without any privacy protection. Our tutor achieves similar performance even after adding noise.

\subsection{Privacy-utility Tradeoff}
\begin{figure}[htbp]
    \centering
    \includegraphics[width=0.45\textwidth]{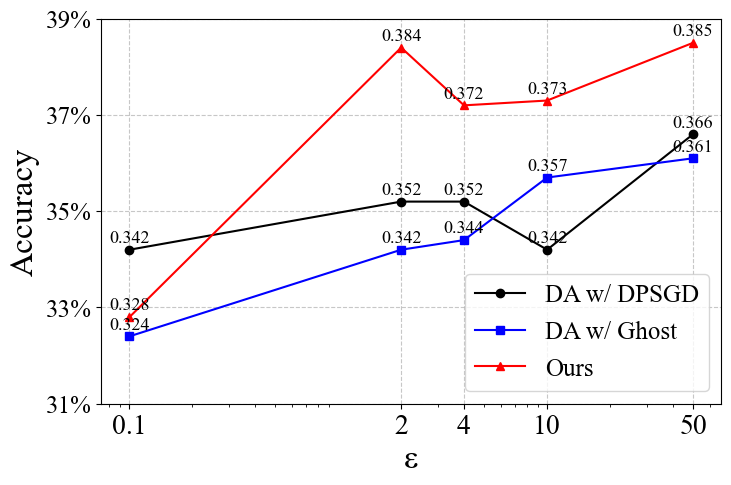}
    \caption{The private-utility tradeoff in accuracy across three DA w/ DP methods at varying $\varepsilon$. The vertical axis represents the accuracy on downstream text classification tasks.}
    \label{fig:eps_acc}
\end{figure}
\begin{figure}[htbp]
    \centering
    \includegraphics[width=0.45\textwidth]{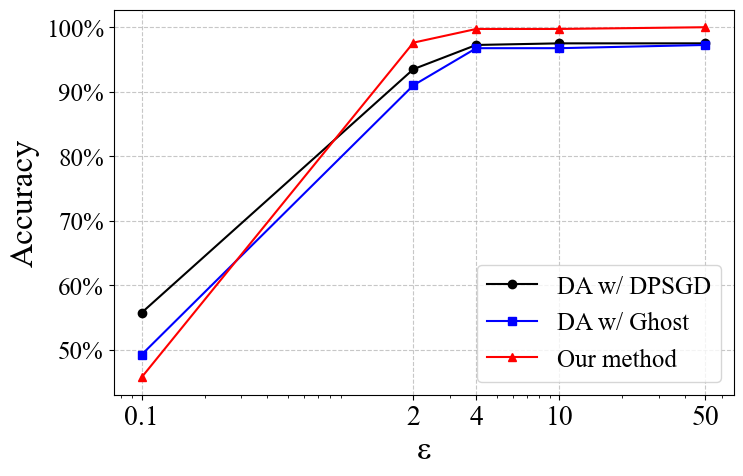}
    \caption{The private-utility tradeoff in the discriminator's accuracy  across three DA w/ DP methods at varying $\varepsilon$. The vertical axis represents the accuracy on our constructed test set.}
    \label{fig:eps_acc_discriminator}
\end{figure}
In Fig.~\ref{fig:eps_acc} and Fig.~\ref{fig:eps_acc_discriminator}, we show the private-utility trade-off curve of three DA w/ DP methods and their discriminators covering the range of meaningful protection (i.e. usually $\varepsilon \in [0.1, 10]$ \cite{triastcyn2020bayesian}). In both figures \texttt{Ours} outperforms \texttt{DA w/ DP-SGD} and \texttt{DA w / Ghost} in most ranges of $\varepsilon \in [0.1, 10]$, showing that our method provides strong privacy protections while having excellent performance.
As $\varepsilon$ increases, privacy protection becomes weaker, leading to a gradual improvement in the performance of various methods. In Fig.~\ref{fig:eps_acc_discriminator}, the Accuracy of the discriminator improves as $\varepsilon$ increases, which aligns with our expectations.
But when $\varepsilon>2$, \texttt{Ours} exhibits fluctuations on downstream tasks. This is because the DP-based discriminator in \texttt{Ours} already achieves near-perfect accuracy ($97.6\%$), and the performance of \texttt{Ours} has exceeded up-bound ($36.7\%$). After exceeding the up-bound, the final performance is probably beyond the control of the discriminator (The discriminator judges whether the synthetic data meet privacy characteristics, but the performance of selected synthetic data by Ours exceeded the original privacy data). 


\subsection{Analysis On Teacher Numbers}

\begin{figure}[htbp]
    \centering
    \includegraphics[width=0.45\textwidth]{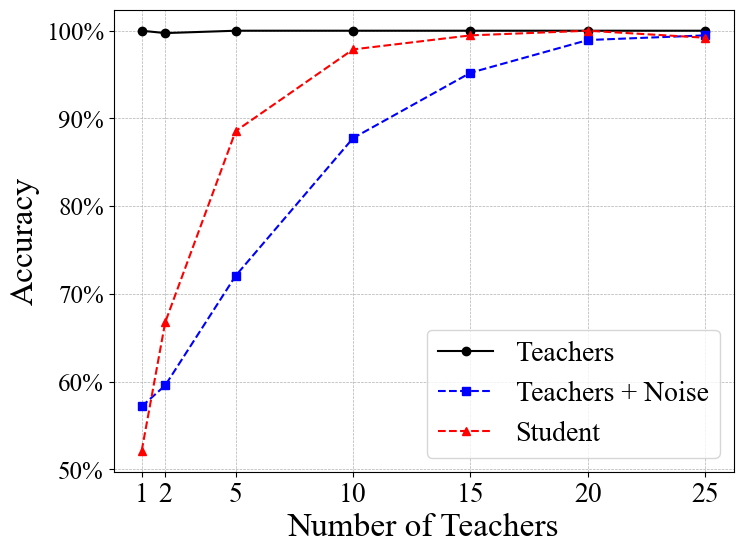}
    \caption{Analysis on teacher number. The vertical axis represents the prediction accuracy of the discriminator on our constructed test set.}
    \label{fig:teacher_num}
\end{figure}

To analyze the impact of teacher number on discriminator, we conducted experiments across varying teacher number with $\varepsilon=4$ (in Fig.~\ref{fig:teacher_num}). \texttt{Teachers} denotes the aggregation of multiple teacher models; \texttt{Teachers + Noise} denotes to add noise to the voting results after aggregating multiple teacher models; \texttt{Student} denotes our final discriminator model. We conclude that: (1) As the teacher number increases,  the performance of \texttt{Teachers + Noise} gradually improves. When the teacher number reaches 20, the impact of adding noise on the aggregated teachers' performance becomes negligible. (2) \texttt{Student}'s performance correlates positively with \texttt{Teachers + Noise}. (3) In most cases, \texttt{Student} outperforms  \texttt{Teachers + Noise}. This is because \texttt{Teachers + Noise}  reduces \texttt{Teachers}' original prediction accuracy by directly adding noise to the inference results. \texttt{Student} trains with noisy data, allowing the model to adapt to the noise.

\subsection{Analysis On Tutor's Distribution}

Fig.~\ref{fig:label_distribution} (See App.~\ref{sec:label_dist}) illustrates two distinct label distributions: \texttt{Private Dist} represents the original private label distribution; \texttt{Tutor Dist} represents the label distribution after adding noise (discussed in §\ref{sec:tutor}), which provides stricter privacy protection by satisfying $\varepsilon= 0.4$. We observe a high consistency between \texttt{Tutor Dist} and \texttt{Private Dist}, which indicates that \texttt{Tutor Dist} effectively retains the characteristics of \texttt{Private Dist}.


\section{Conclusion}
In this paper, we proposed a DP-based DA method for text classification in private domains, which prompts a LLM to generate pseudo samples and uses a DP-based discriminator to examine the LLM's outputs. In this way, we transfer pseudo text generation task, which is a challenging DP paradigm, to a discrimination task. We construct a DP-based discriminator via knowledge distillation and construct a DP-based tutor to guide the sample generation with a low privacy cost. Theoretical analyses illustrates the bound of protections of our models and our experimental results shows our model's effectiveness.

\section{Limitations}

In our study, there were several limitations.

(1) First of all, we use GPT-3.5 instead of GPT-4 in our experiments and GPT-3.5 is not a latest and SOTA GPT API, which seems to limited the performance of our model. 
The main reason is that our baselines \cite{dpsgd2016deep,li2021large} are all based on GPT-3.5 for a fair comparison. In addition, compared to GPT-3.5, GPT-4 costs too much to obtain API keys. According to the official website, the fee for GPT 3.5 is 0.002\$/1k Token, and the fee for GPT 4 is 0.06\$/1k Token. In our actual baseline comparison, about 560,000 pieces of data were generated using GPT-3.5, and the total cost was about 300\$. And if GPT-4 is used, the same number of tokens will cost more than 9000\$. Therefore, the use of GPT-3.5 can not only effectively reduce the cost significantly, but also there will not be much difference in the generation effect. At present, we use GPT-3.5 for method verification, and GPT-4 will be used for effect verification in the future.

(2) Secondly, our approach is based on LLM-generated data, relatively dependent on the quality of the generated text. If the data are difficult to generate, or the overall quality of the generated text is poor, this may limit the advantages of our approach. Subsequently, we can leverage approaches such as prompt tuning, by carrying out a variety of excellent prompt engineering during the training phase of the LLM, enhancing the quality of text generated by LLM.

\section{Ethical Considerations}
We place significant importance on ethical considerations and adhere rigorously to the ACL Ethics Policy.

(1) This study introduces a novel text classification method, utilizing an LLM to generate pseudo samples and a DP-based discriminator to evaluate them, without ethical concerns regarding motivation or algorithmic approach, as no private information is used.

(2) Nevertheless, it's crucial to contemplate scenarios where individuals deliberately exploit our model for illicit purposes. For instance, someone might use the text generation model, used in this paper for generating pseudo data, to fabricate fake text or misinformation. This potential misuse poses a negative societal impact, using our model to generate false medical reports. Moving forward, we intend to implement constraints within our model to prevent the generation of texts for illegal activities, such as introducing filters to identify and flag potentially harmful or illegal content.

(3) Moreover, it's imperative to exercise caution in utilizing our model and refrain from assuming its infallibility. One potential unethical application involves gathering data from users who believe our model guarantees complete privacy protection, potentially overlooking the actual strength of privacy safeguards. This oversight could lead to adverse societal consequences. Therefore, we urge researchers intending to utilize this model to prioritize the efficacy of privacy protection. Additionally, measures should be taken to prevent researchers from collecting data from users who lack a proper understanding of our algorithm. For example, Clarify privacy policies and rules for data usage, and restrict access to collected data to authorized personnel only. We recommend that researchers ensure users contributing their data comprehend the risks associated with our model fully.

(4) In general, if the dataset contains privacy, it may be leaked during use. Regarding the datasets utilized in our experiments, the dataset generated by LLM does not contain the personally identifiable information of the real user. In contrast, the Medical Transcriptions dataset contains sample medical transcriptions for various medical specialties, allowing us to conduct experiments to assess the efficacy of privacy protection measures. It's important to note that the Medical Transcriptions dataset was previously made available to the public. Therefore, our research in this paper does not involve releasing any additional personal information of users.
\bibliography{custom}

\appendix
\section{Deduction of Sensitivity $\Delta_{KD}$}
\label{sec:sensitivity_kd}
We obtain the Equations (\ref{eq:sensitivity_kd}) in §\ref{sec:privacy_analyses} of the paper body since $P_{i}(\mathcal{Y}|x)$ and $P_i^\prime(\mathcal{Y}|x)$ are the output distribution of $i$-th teacher model. where $x$ is a given sample and $\mathcal{Y}=\{0,1\}$.
\begin{align}
    \Delta_{KD} &= \| h(\mathcal{D}) - h(\mathcal{D'}) \|_2 \\
    &\le \| P_i(\mathcal{Y}|x) - P_i^\prime(\mathcal{Y}|x) \|_2 \nonumber \\
    &=\bigg(\sum_{j=1}^{|\mathcal{Y}|} (P_{ij}(\mathcal{Y}|x) - P_{ij}^{\prime}(\mathcal{Y}|x))^2 \bigg)^{1/2} \nonumber
\end{align}

We know $(P_{ij}(\mathcal{Y}|x) - P_{ij}^{\prime}(\mathcal{Y}|x))^2$ is smaller than $|P_{ij}(\mathcal{Y}|x) - P_{ij}^{\prime}(\mathcal{Y}|x)|$ since $|P_{ij}(\mathcal{Y}|x) - P_{ij}^{\prime}(\mathcal{Y}|x)| \in (0, 1)$ for each $j$. Hence, we have,
\begin{align}
   &\bigg(\sum_{j=1}^{|\mathcal{Y}|} (P_{ij}(\mathcal{Y}|x) - P_{ij}^{\prime}(\mathcal{Y}|x))^2 \bigg)^{1/2} \\
   \le &\bigg(\sum_{j=1}^{|\mathcal{Y}|} |P_{ij}(\mathcal{Y}|x) - P_{ij}^{\prime}(\mathcal{Y}|x)| \bigg)^{1/2} \nonumber \\ 
   \le &\bigg(\sum_{j=1}^{|\mathcal{Y}|} |P_{ij}(\mathcal{Y}|x) + P_{ij}^{\prime}(\mathcal{Y}|x)| \bigg)^{1/2} \nonumber
\end{align}

We know $|a+b| = a + b$ when $a,b \in (0,1)$, so we have,
\begin{align}
    &\bigg(\sum_{j=1}^{|\mathcal{Y}|} |P_{ij}(\mathcal{Y}|x) + P_{ij}^{\prime}(\mathcal{Y}|x)| \bigg)^{1/2} \\
    = &\bigg(\sum_{j=1}^{|\mathcal{Y}|} P_{ij}(\mathcal{Y}|x) + \sum_{j=1}^{|\mathcal{Y}|} P_{ij}^{\prime}(\mathcal{Y}|x) \bigg)^{1/2} \nonumber \\
    = &\bigg(1 + 1 \bigg)^{1/2} \le \sqrt{2}, \nonumber 
\end{align}

In summary, the upper bound of the sensitivity is,
\begin{align}
    \Delta_{KD} &= \| h(\mathcal{D}) - h(\mathcal{D'}) \|_2 \\
    &\le \| P_i(\mathcal{Y}|x) - P_i^\prime(\mathcal{Y}|x) \|_2 = \sqrt{2}. \nonumber
\end{align}

\begin{figure*}[h]
    \centering
    \includegraphics[width=1\textwidth]{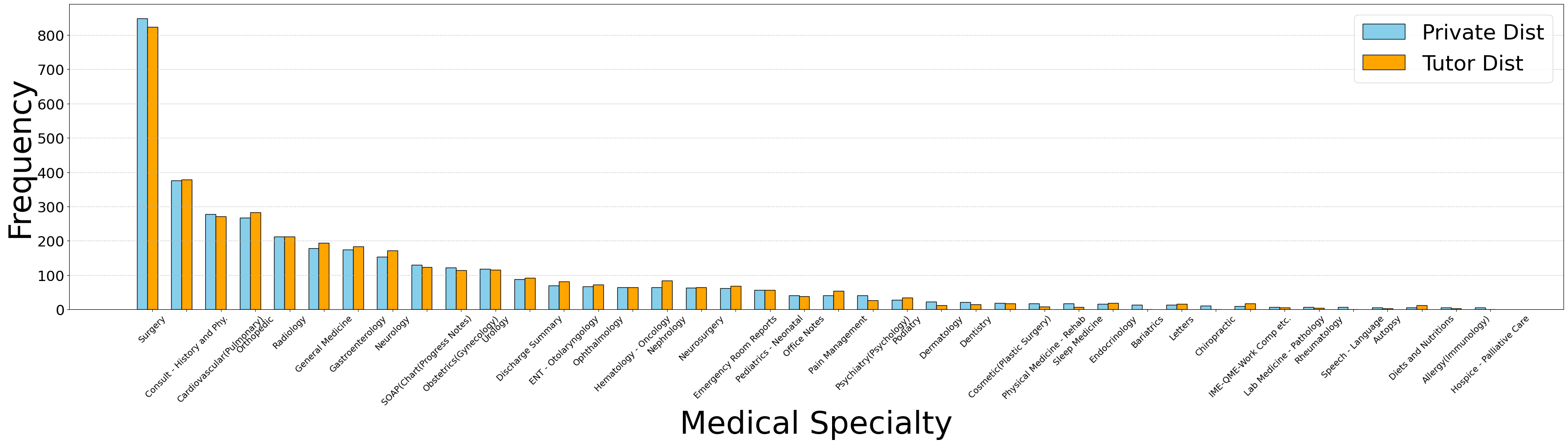}
    \caption{Label distributions. The horizontal axis enumerates all data labels, while the vertical axis represents the frequency of the labels.}
    \label{fig:label_distribution}
\end{figure*}

\section{Deduction of Sensitivity $\Delta_{tutor}$}
\label{app:sensitivity_tutor}
We obtain the Equations (\ref{eq:sensitivity_tutor}) in §\ref{sec:privacy_analyses} of the paper body since $P_{l}(\mathcal{C})$ and $P^\prime_{l}(\mathcal{C})$ are the sample $l$'s distribution. 
\begin{align}
    \Delta_{tutor} &= \| g(\mathcal{D}) - g(\mathcal{D'}) \|_2 \\
     &\le \| P_l(\mathcal{C}) - P_l^\prime(\mathcal{C}) \|_2 \nonumber \\
    &=\bigg(\sum_{v=1}^{|\mathcal{C}|} (P_{lv}(\mathcal{C}) - P_{lv}^\prime(\mathcal{C}))^2 \bigg)^{1/2} \nonumber
\end{align}

Because $l$'s distribution is a binary probability distribution with values 1 on its labels and 0 on all other classes, $\sum_{v=1}^{|\mathcal{C}|} (P_{lv}(\mathcal{C}) - P_{lv}^\prime(\mathcal{C}))^2$ has a maximum value when sample $l$ belongs to different categories in neighboring datasets $\mathcal{D}$ and $\mathcal{D}^\prime$.
\begin{align}
    &\bigg(\sum_{v=1}^{|\mathcal{C}|} (P_{lv}(\mathcal{C}) - P_{lv}^\prime(\mathcal{C}))^2 \bigg)^{1/2} \\
    \le&\bigg(1 + 1 \bigg)^{1/2} = \sqrt{2} \nonumber 
\end{align}

In summary, the upper bound of the sensitivity is,
\begin{align}
    \Delta_{tutor} &= \| g(\mathcal{D}) - g(\mathcal{D'}) \|_2 \\
    &\le \| P_l(\mathcal{C}) - P_l^\prime(\mathcal{C}) \|_2 \le \sqrt{2}. \nonumber
\end{align}

\section{Detailed Deduction of Composition (KD and Tutor)}
\label{app:composition}
The KD algorithm $\mathcal{M}_{KD}$ satisfies ($\varepsilon_{KD}$, $\delta_{KD}$)-DP and the tutor algorithm $\mathcal{M}_{tutor}$ satisfies ($\varepsilon_{tutor}$, $\delta_{tutor}$)-DP. Since the output results of KD and Tutor are independent, for $\forall{(\mathcal{D}_{KD} \times \mathcal{D}_{tutor}) \in (\mathcal{R}_{KD} \times \mathcal{R}_{tutor})}$, there is
\begin{align}
    &\Pr[(\mathcal{M}_{KD}),(\mathcal{M}_{tutor}) \in (\mathcal{D}_{KD} \times \mathcal{D}_{tutor})] \nonumber \\ 
    =&\Pr[(\mathcal{M}_{KD}) \in \mathcal{D}_{KD}]\Pr[(\mathcal{M}_{tutor}) \in \mathcal{D}_{tutor}] \nonumber \\
    \le &([e^{\varepsilon_{tutor}}Pr^\prime[(\mathcal{M}_{tutor}) \in \mathcal{D}_{tutor}]] \wedge 1 +\delta_{tutor}) \nonumber \\
    &\times\Pr[(\mathcal{M}_{KD}) \in \mathcal{D}_{KD}] \nonumber \\
    \le &([e^{\varepsilon_{tutor}}Pr^\prime[(\mathcal{M}_{tutor}) \in \mathcal{D}_{tutor}]] \wedge 1) \nonumber \\
    &\times\Pr[(\mathcal{M}_{KD}) \in \mathcal{D}_{KD}] \nonumber \\&+\delta_{tutor}\Pr[(\mathcal{M}_{KD}) \in \mathcal{D}_{KD}] \nonumber \\
    \le &e^{\varepsilon_{tutor}}Pr^\prime[(\mathcal{M}_{tutor}) \in \mathcal{D}_{tutor}] \nonumber \\
    &\times Pr[(\mathcal{M}_{KD}) \in \mathcal{D}_{KD}]+\delta_{tutor} \nonumber \\
    \le &e^{\varepsilon_{KD+tutor}}Pr^\prime[(\mathcal{M}_{tutor}) \in \mathcal{D}_{tutor}] \nonumber \\
    &\times [Pr^\prime[(\mathcal{M}_{KD}) \in \mathcal{D}_{KD}]+\delta_{KD}]+\delta_{tutor} \nonumber \\
    \le &e^{\varepsilon_{KD+tutor}}Pr^\prime[(\mathcal{M}_{tutor}) \in \mathcal{D}_{tutor}] \nonumber \\
    &\times Pr^\prime[(\mathcal{M}_{KD}) \in \mathcal{D}_{KD}]+\delta_{KD}+\delta_{tutor} \nonumber \\
    =& e^{\varepsilon_{KD+tutor}}  \nonumber \\
    &\times \Pr^\prime[(\mathcal{M}_{KD}),(\mathcal{M}_{tutor}) \in (\mathcal{D}_{KD} \times \mathcal{D}_{tutor})] \nonumber \\
    &+\delta_{KD}+\delta_{tutor}.
\end{align}
Therefore, it can be concluded from the above derivation that
the combination $\mathcal{M}$ satisfies ($\varepsilon_{KD} + \varepsilon_{tutor}$, $\delta_{KD} + \delta_{tutor}$)-DP.

\section{Implementation Details}
\label{sec:implementation details}
We select ``gpt-3.5-turbo'' with maximum 4,096 tokens construct synthetic text. 
In the DP-based DA. We fine-tune teachers and students based on the public pre-trained RoBERTa-large\footnote{We have tried LLM (i.e. GPT-3.5) as  discriminator with the accuracy is less than 60\%.} \cite{bert2019naacl_devlin}, which is representative and performs well in binary classification tasks. We set default number of teacher models is 15 and use the AdamW \cite{loshchilov2017decoupled} optimizer with a learning rate of $2\times 10^{-5}$. The default target $\varepsilon$ is 4 in §\ref{sec:discriminator} and 0.4 in §\ref{sec:tutor}, both $\delta$ is $10^{-6}$. We use the autoDP\footnote{https://github.com/yuxiangw/autodp} to obtain the standard deviation $\sigma$ of the Gaussian mechanism \cite{dong2019gaussian} is 6. 
In the downstream classification task, we use ClinicalBERT\cite{huang2019clinicalbert} as the PLM with batch size of 64 and the AdamW optimizer with the learning rate of $10^{-6}$. ClinicalBERT is a BERT derivative specifically for clinical medicine, outperforms RoBERTa-large in the medical field.
All experiments were performed using a  single NVIDIA RTX 3090 GPU.



\section{Label Distributions}
\label{sec:label_dist}

Fig.~\ref{fig:label_distribution} plots two distinct label distributions: \texttt{Private Dist} represents the original private label distribution; \texttt{Tutor Dist} represents the label distribution of tutor. 

\end{document}